\documentclass[lettersize,journal]{IEEEtran}
\usepackage{graphicx}
\usepackage{multirow}
\usepackage{subfigure}
\usepackage{ifpdf}
\usepackage{diagbox}
\usepackage{bbm}
\usepackage{amssymb}
\usepackage{booktabs}
\usepackage{makecell}
\usepackage{colortbl}  
\usepackage{xcolor}
\usepackage{float}
\usepackage{acronym}

\usepackage{colortbl}  
\definecolor{pinkkkk}{RGB}{255,102,255}
\definecolor{pinkkk}{RGB}{237,218,244}
\definecolor{pinkkk1}{RGB}{248,235,249}
\definecolor{redd}{RGB}{205,143,143}
\definecolor{grayyy}{RGB}{242,242,242}
\definecolor{greenn}{RGB}{124,191,51}
\usepackage{pifont}
\usepackage{algpseudocode}
\usepackage{amsmath,amsfonts}
\usepackage{algorithm}
\usepackage{array}
\usepackage[caption=false,font=normalsize,labelfont=sf,textfont=sf]{subfig}
\usepackage{textcomp}
\usepackage{stfloats}
\usepackage{url}
\usepackage{verbatim}
\usepackage{cite}

\begin{document}
\title{MMO-IG: Multi-Class and Multi-Scale Object Image Generation for Remote Sensing}

\author{Chuang Yang,
		Bingxuan Zhao,
		Qing Zhou,
		Qi Wang*,~\IEEEmembership{IEEE Member}
\thanks{
	This work was supported by the National Natural Science Foundation of China under Grant 62471394 and U21B2041.

Chuang Yang and Qi~Wang are with the School of Artificial Intelligence,  OPtics and ElectroNics (iOPEN), Northwestern Polytechnical University, Xi'an 710072, Shaanxi, P. R. China. 
	
Bingxuan~Zhao and Qing Zhou are with the School of Computer Science, and with the School of Artificial Intelligence, OPtics and ElectroNics (iOPEN), Northwestern Polytechnical University, Xi'an 710072, Shaanxi, P. R. China. 
	
E-mail: omtcyang@gamil.com, bxuanzhao202@gmail.com, chautsing@gmail.com, crabwq@gmail.com.}
\thanks{Chuang Yang and Bingxuan Zhao contributed equally to this work.}
\thanks{Qi Wang is the corresponding author.}

}

\markboth{}%
{Shell \MakeLowercase{\textit{et al.}}: A Sample Article Using IEEEtran.cls for IEEE Journals}


\maketitle

\begin{abstract}
The rapid advancement of deep generative models (DGMs) has significantly advanced research in computer vision, providing a cost-effective alternative to acquiring vast quantities of expensive imagery. However, existing methods predominantly focus on synthesizing remote sensing (RS) images aligned with real images in a global layout view, which limits their applicability in RS image object detection (RSIOD) research. To address these challenges, we propose a multi-class and multi-scale object image generator based on DGMs, termed \textit{MMO-IG}, designed to generate RS images with supervised object labels from global and local aspects simultaneously. Specifically, from the local view, MMO-IG encodes various RS instances using an iso-spacing instance map (ISIM). During the generation process, it decodes each instance region with iso-spacing value in ISIM—corresponding to both background and foreground instances—to produce RS images through the denoising process of diffusion models. Considering the complex interdependencies among MMOs, we construct a spatial-cross dependency knowledge graph (SCDKG). This ensures a realistic and reliable multidirectional distribution among MMOs for region embedding, thereby reducing the discrepancy between source and target domains. Besides, we propose a structured object distribution instruction (SODI) to guide the generation of synthesized RS image content from a global aspect with SCDKG-based ISIM together. Extensive experimental results demonstrate that our MMO-IG exhibits superior generation capabilities for RS images with dense MMO-supervised labels, and RS detectors pre-trained with MMO-IG show excellent performance on real-world datasets. Code is available at \textcolor{pinkkkk} {\url{https://github.com/omtcyang/MMO-IG}}.
\end{abstract}

\begin{IEEEkeywords}
Remote sensing, image generation, object detection, diffusion model.
\end{IEEEkeywords}

\begin{figure}[t]
	\centering
	\includegraphics[width=0.99\linewidth]{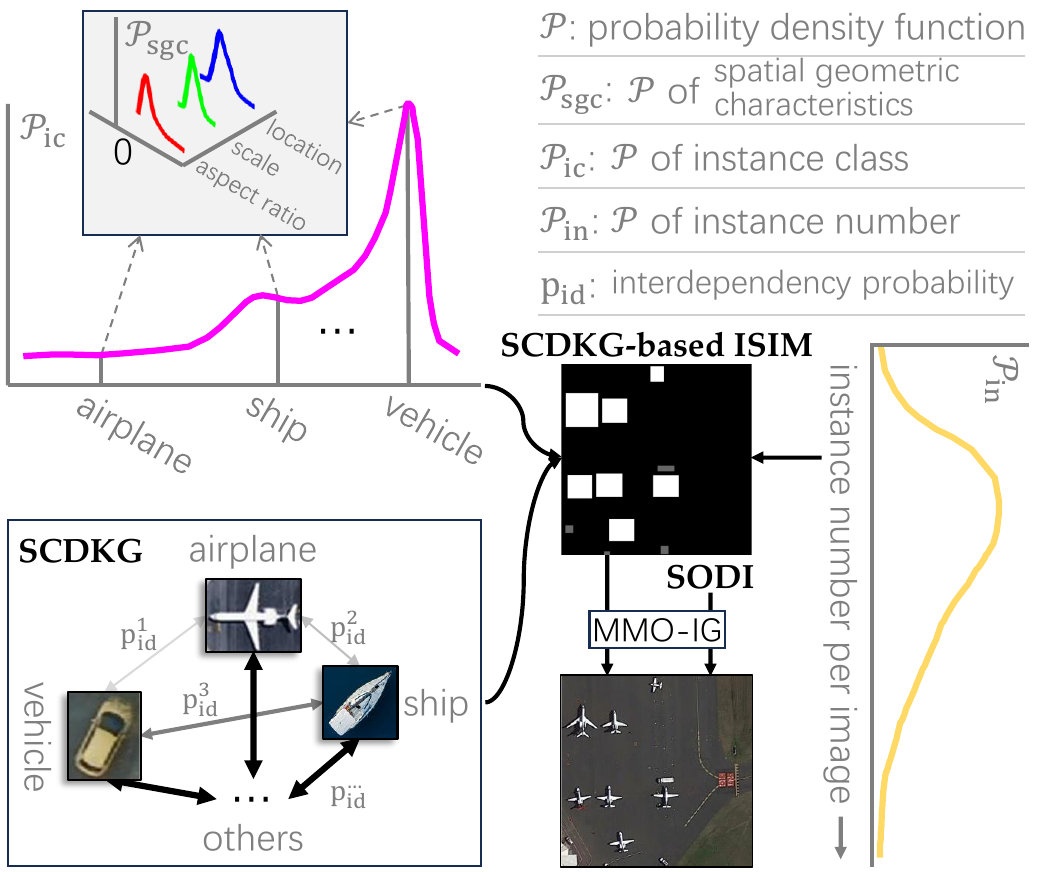}
	\caption{Illustration of the generation of RS images containing MMOs by the proposed MMO-IG. Notably, each RS object is modeled by a unique $\mathcal{P}_{\rm sgc}$ according to the corresponding realistic geometric characteristics.}
	\label{fig:V1}
\end{figure}

\section{Introduction}
\IEEEPARstart{O}{bject} detection in remote sensing (RS) images~\cite{li2023instance,zhang2023efficient,gao2023global,li2023large} is an important task in earth observation technology, providing essential support for various application scenarios, such as disaster monitoring, military reconnaissance, traffic monitoring, and smart cities. This task aims to determine object categories and locations based on input optical RS images. With the progress of deep learning-based generic detection technology~\cite{ren2016faster,yang2022cm,girshick2014rich,yang2023text}, remote sensing image object detection (RSIOD) research has achieved significant development, where analyzing complex backgrounds and focusing on varied class and scale foreground objects simultaneously is becoming achievable as long as the algorithm that is well-trained based on sufficient and high-quality data. However, the high cost of acquiring satellite RS images and the labor-intensive nature of image annotation limit the availability of data for this research area, making it challenging to adequately train algorithms. This widespread data limitation issue in the field of RS has consistently hindered researchers from making further breakthroughs. 

Recently, the rapidly developed deep generative models (DGMs)~\cite{ho2020denoising,rombach2022high,zhang2023adding} provide a solution for alleviating the above issue. They aim to generate realistic and high-quality images. It enhances the accessibility of data in resource-constrained environments significantly, which makes it possible to build RS generative models (RSGMs)~\cite{tang2024crs,qian2020generating} for alleviating the data limitation of RSIOD. While existing RSGMs follow generic DGMs to focus on the achievement of text-to-image or layout-to-image~\cite{li2023gligen,zheng2023layoutdiffusion,yang2023reco,sun2019image}, where the former generates RS images according to text descriptions prompt and the latter is designed to render RS images based on a given layout image prompt (e.g., depth map, sketch, road, edge, etc.). Traditional approaches relying on text or layout prompts face three fundamental limitations when handling Multi-Class and Multi-Scale Objects (MMOs): (1) Layout prompts struggle to encode instance-specific scale variations and precise spatial relationships, often producing rigid object arrangements inconsistent with real RS scenes; (2) Existing methods prioritize global prompt-image alignment while neglecting local interdependencies between adjacent objects; (3) Fixed spatial constraints in layout guidance limit flexible adaptation to diverse RS observation geometries. It means they are hard to provide qualified data with instance-level labels that can model complex interdependencies among different RS objects for the RSIOD task. 

Following the above issue, we introduce an instance-aware image generator for providing sufficient and high-quality data with multi-class and multi-scale object (MMO) supervision information for RS detection, namely MMO-IG. To overcome the limitations of traditional guidance signals, we propose the Iso-spacing Instance Map (ISIM) -- a novel control mechanism where distinct grayscale regions with unique iso-spacing values encode different object classes. This approach preserves three critical instance attributes: (1) Spatial coordinates through region centroids, (2) Object scales via region areas, and (3) Class identities by grayscale intensities, enabling simultaneous control of multiple object instances while maintaining flexible placement. Specifically, inspired by popular diffusion-based controlled generative models~\cite{zhang2023adding,zhao2024uni,li2025controlnet,zavadski2023controlnet}, we formulate the MMO-IG as a controllable generation problem. Different from previous methods that were designed to generate RS images whose content adheres to global layout requirements, the proposed MMO-IG is demanded to focus on each object instance's class, location, and scale in a local view while considering the interdependencies among different instances from a global aspect. Considering that the existing text and layout prompt cannot represent MMOs, an iso-spacing instance map (ISIM) is introduced as the control signal for guidance in our model in generating RS images containing MMPs, where instances with different classes are encoded via various iso-spacing valued grayscale regions. During testing, MMO-IG decodes these regions into instances with different classes according to regions' grayscale values while keeping the location and scale characteristics of the corresponding regions on generated instances. 

Existing RSGMs predominantly focus on pairwise object relationships, failing to capture the complex network of dependencies in real RS scenes where objects form hierarchical clusters (e.g., vehicle groups around buildings) and exhibit cross-class interactions (e.g., ships near ports). To address this, we develop the Spatial-Cross Dependency Knowledge Graph (SCDKG) that explicitly models four types of spatial relationships: (1) Intra-class proximity constraints, (2) Cross-class attraction/repulsion forces, (3) Orientation-aware co-occurrence patterns, and (4) Scale compatibility rules.Besides, distinct from previous RSGMs concentrate on pursuing a correct corresponding between the given text or layout prompt and the generated RS image, there exist complex interdependencies among different objects with the same class and also in different classes of objects, which demands our model to ensure a rational interdependent distribution among the MMOs on the generated RS image. Based on this consideration, we construct a spatial-cross dependency knowledge graph (SCDKG) to formulate the complex interdependencies among different RS objects. During testing, we synthesize an ISIM with rational interdependent distribution among the MMOs under the guidance of SCDKG to render the RS image via MMO-IG more accurately reflective of reality. Based on the above proposed ISIM and SCDKG, we construct MMO-IG for generating mass RS images with dense instance-level labels, which provides a solution for alleviating the data limitation problem that exists in the RSIOD field. Furthermore, the hallucination problem in deep generative models (DGMs) can cause remote sensing (RS) objects to appear incorrectly in the background, leading to discrepancies between instance-level labels (i.e., ISIM) and the generated RS images, which negatively impacts downstream tasks. To address this issue, SODI is introduced to guide the image generation process by considering the global perspective of image style and instance characteristics, ensuring both accuracy and control.

In summary, the contributions of our work are fourfold:
\begin{enumerate}
	\item An iso-spacing instance map (ISIM) is designed as the control signal for guidance in generating RS images filled with dense MMOs. ISIM is an effective and intuitive encoding method for RS objects with different classes, locations, and scales, which helps the model comprehend objects' geometric characteristics for generating high-quality data.
	
	\item A spatial-cross dependency knowledge graph (SCDKG) is built, which formulates the complex interdependencies among different RS objects for providing support in synthesizing ISIM with a rational spatial distribution for the MMOs. It helps render the RS image more accurately reflective of reality.
	
	\item A SODI is introduced to ensure the alignment between ISIM and the generated RS image. It assists in regulating image content to align with the remote sensing style. Meanwhile, the object statistics description ensures strict correspondences between MMOs in ISIM and the generated remote sensing image, which facilitates an accurate and controllable image generation process.
	
	\item An ISIM and SCDKG-based RS image generation model, termed MMO-IG, is developed to decode synthetic SCDKG-guided ISIM with dense instance regions into RS images through a denoising process. This model provides abundant training data for RSIOD, helping to alleviate existing data limitations such as scarcity and sample imbalance.
\end{enumerate}

The remainder of the paper is structured as follows: Section~\ref{Related Work} reviews the contributions of scholars in related fields, specifically DGMs, and RSGMs, and evaluates the strengths and limitations of existing algorithms, which inform the design of MMO-IG. Section~\ref{Methodology} presents a detailed visualization of the proposed MMO-IG. Section~\ref{Experiments} discusses the results of ablation studies and comparative analyses. Finally, Section~\ref{Conclusion} summarizes the paper's contributions to the data limitation problem of RSIOD.

\section{Related Work}
\label{Related Work}
The rapid development of DGMs progress RSGMs significantly. In this section, we introduce the related works in the research of DGMs and RSGMs briefly.

\subsection{Remote Sensing Image Object Detection}
Object detection is a key topic in computer vision, which provides sufficient support to analyze image context. Inspired by generic object detection approaches~\cite{redmon2018yolov3,yang2022bip,liu2018path,xiao2023corporate,law2018cornernet,yang2023instance}, researchers designed RSIOD frameworks according to the specific geometric characteristics of RS objects, which become critical techniques for RS scenario applications. Existing RSIOD methods can be classified into small~\cite{li2020cross,gao2023global,zhang2023superyolo} and oriented~\cite{zhang2023efficient,qian2023building,li2023feature} object detection methods roughly according to the object characteristics. The former concentrated on designing the feature fusion structure~\cite{li2025tbnet,xiaolin2022small,ma2022feature} or introducing attention mechanism into frameworks~\cite{ye2022adaptive,dong2022attention,gong2022swin} to obtain strong expression ability to help the model distinguish small targets from complex backgrounds more effectively. The latter primarily aimed to represent oriented object boundaries~\cite{fu2020rotation,yang2021dense} accurately to avoid superfluous background interference. Besides, some works~\cite{DBLP:journals/tgrs/LiCWZH23,DBLP:journals/lgrs/LinZWT23,DBLP:journals/tgrs/PangLSXF19} focus on optimizing the whole detection framework to achieve performance balance between accuracy and efficiency. Although these methods achieve competitive performance on multiple public datasets, the limited availability of data in many scenarios hinders their ability to maintain superior performance.

\subsection{Deep Generative Models}
Deep generative models (DGMs) are designed for creating realistic synthetic data and enhancing data-driven applications across various fields. Initially, variational auto-encoders (VAE)~\cite{DBLP:journals/corr/KingmaW13,DBLP:conf/nips/PuGHYLSC16,DBLP:conf/icml/KusnerPH17} and generative adversarial networks (GAN)~\cite{DBLP:conf/nips/GoodfellowPMXWOCB14,DBLP:conf/nips/LiuT16,DBLP:conf/iccv/MaoLXLWS17} advanced DGM research considerably and had gained increasing influence in computer vision tasks. However, VAEs often produce blurry outputs due to their reliance on Gaussian assumptions, while GANs can be challenging to train and are prone to mode collapse. In contrast, diffusion models~\cite{sohl2015deep,ho2020denoising,rombach2022high,zhang2023adding} addressed these limitations by employing a stepwise noise reduction process that yields sharper and more stable outputs, capturing researchers' attention. Specifically, DDPM~\cite{ho2020denoising} built on foundational concepts introduced in the original diffusion model~\cite{sohl2015deep} by refining the diffusion process to improve sample quality and stability while maintaining flexibility for various tasks. Stable Diffusion~\cite{rombach2022high} introduced a latent space to optimize computational efficiency and accelerate image generation processes. However, these methods lack advanced control mechanisms, limiting their ability to precisely guide image generation. Addressing these issues, ControlNet~\cite{zhang2023adding} enhanced generated outputs by integrating additional neural networks that conditionally guide the generation process based on specific input features (e.g., canny edge, human pose, and sketch, etc.), which provided an effective solution for alleviating the data limitation problem exists in remote sensing image analysis.
\begin{figure*}[t]
	\centering
	\includegraphics[width=0.99\linewidth]{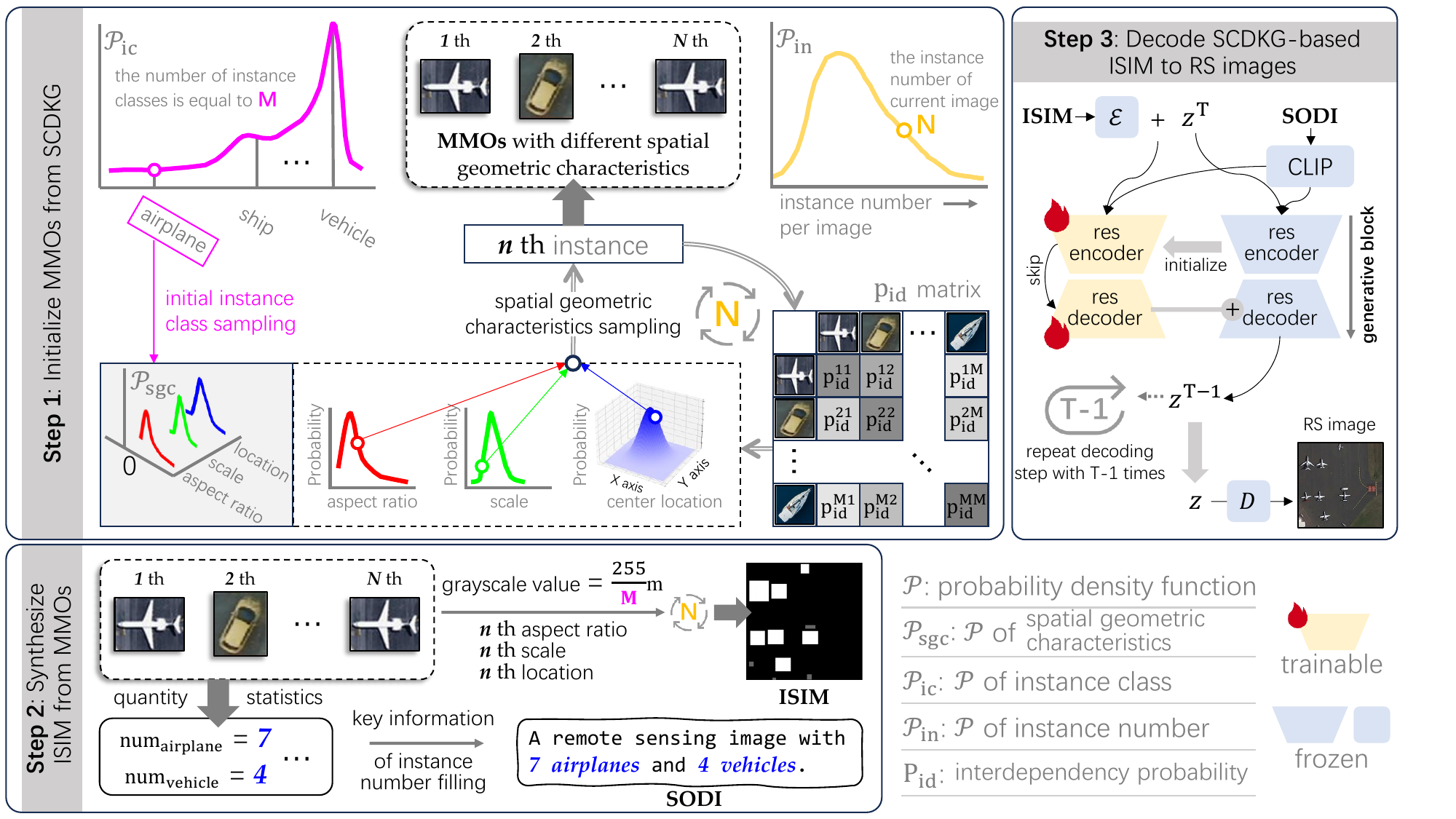}
	\caption{Overall pipeline of MMO-IG for generating RS images with dense instance-level bounding box labels.  It first synthesizes rational spatial geometric characteristics of MMOs via SCDKG. They then are encoded via the designed ISIM while describing the RS image content through SODI.  In the end, following the diffusion model to decode the ISIM to RS image contained MMOs under the guidance of SODI. Notably, each RS object is modeled by a unique $\mathcal{P}_{\rm sgc}$ according to the corresponding realistic geometric characteristics.}
	\label{fig:V2}
\end{figure*}
\subsection{Remote Sensing Generative Models}
In remote sensing image analysis research, data acquisition is particularly challenging compared to other computer vision fields, leading to a more severe issue of data scarcity. Based on the above consideration, researchers introduce DGMs into remote sensing image analysis. The authors~\cite{qian2020generating} generate multiple sets of pseudo-labeled samples from real data by SinGAN~\cite{shaham2019singan} while a novel quantitative sifting metric is then applied to assess the authenticity and diversity of these samples, enabling the selection of the most optimal pseudo-labeled samples for enhancing model training. To achieve image directional generation, D-SGAN~\cite{lv2021remote} takes a rough segmentation map as input to guide the generation process. RSDiff~\cite{sebaq2024rsdiff} leverages diffusion models to generate remote sensing images from textual descriptions, enhancing the synthesis quality and semantic alignment between text and imagery. CRS-Diff~\cite{tang2024crs} enables precise manipulation of image attributes through guided compound control conditions, such as the integration of text, depth maps, sketches, and road layouts, thereby enhancing the controllability and specificity of remote sensing image generation. Existing RSGMs pursue text-to-image or layout-to-image based on DDPM~\cite{ho2020denoising} or GAN~\cite{DBLP:conf/nips/GoodfellowPMXWOCB14} and achieve superior performance. However, a generation model enables provide sufficient data with instance-level labels for the RSIOD task, which is still under-explored. 
 
\section{Methodology}
\label{Methodology}
RSIOD is an important task for plenty of applications (e.g., disaster monitoring, military reconnaissance, traffic monitoring, and smart cities) in the field of remote sensing image analysis. To alleviate the data limitation problem in RSIOD research, we propose MMO-IG in this paper, the method details will be described in the following paragraphs.

\subsection{Overall Pipeline}
An image generation method, namely MMO-IG, tailored for the RSIOD task has been proposed. The overall pipeline of MMO-IG is visualized in Fig.~\ref{fig:V2}. It can be found that the whole generation process consists of three steps: 1) initializing MMOs from SCDKG; 2) synthesizing ISIM and SODI from MMOs; 3) decoding SCDKG-based ISIM to the RS image. On the whole, MMO-IG decodes the regions with different grayscale values in ISIM to instances with different classes while keeping the regions' spatial geometric characteristics (i.e., aspect ratios, scales, and locations). Meanwhile, it ensures the whole image style under the guidance of SODI. Considering the complex spatial-cross dependency relationships among dense objects, SCDKG is designed to ensure a rational instance-level layout that corresponds to realistic scenes. In the following subsections, we will describe SCDKG, ISIM, and SODI in detail respectively.
\begin{figure}[t]
	\centering
	\includegraphics[width=0.99\linewidth]{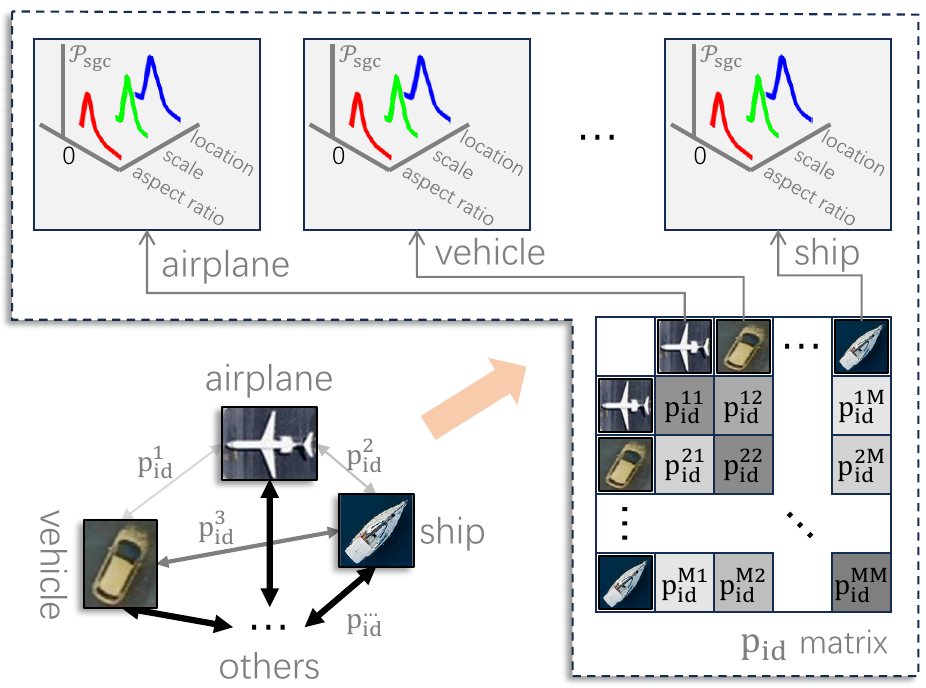}
	\caption{Illustration of the proposed SCDKG, which models complex interdependencies among objects of different classes via $\rm p_{id}$ matrix and their diverse spatial geometric characteristics via $\mathcal{P}_{\rm sgc}$. Notably, each RS object is modeled by a unique $\mathcal{P}_{\rm sgc}$ according to the corresponding realistic geometric characteristics.}
	\label{fig:V3}
\end{figure}
\subsection{Spatial-Cross Dependency Knowledge Graph}

As previously introduced, the proposed MMO-IG aims to decode regions with varying grayscale values in ISIM into RS objects. Unlike generic DGMs that focus on realistic salient object details or existing RSGMs that align layout-controlled conditions with corresponding ground truth, MMOs distributed on RS images require MMO-IG to address complex interdependencies among objects of different classes and their diverse spatial geometric characteristics (as illustrated in Fig.~\ref{fig:V3}). This approach ensures a rational and realistic content distribution for the generated RS images. In light of these considerations, SCDKG is designed to model the intricate spatial relationships among RS objects.

SCDKG is tasked with determining MMOs with distinct spatial geometric characteristics for synthesizing ISIM (the corresponding procedures are detailed in Step~1 of Fig.\ref{fig:V2} and Algorithm~\ref{alg:A1}). Specifically, SCDKG initially samples the RS object class based on the instance class probability density function $\mathcal{P}_{\rm ic}$, which describes the distribution of RS objects across various types within a given dataset.
\begin{algorithm}[t]
	\label{algorithm1}
	\caption{Initialize MMOs from SCDKG}  
	\begin{algorithmic}
		\Require The probability density functions $\mathcal{P}_{\rm ic}$, $\mathcal{P}_{\rm in}$, and SCDKG ($\rm{p}_{\rm id}$ matrix and $\mathcal{P}_{\rm sgc}$);
		\Ensure object list $\rm L_{obj}$ of MMOs for synthesizing ISIM;
		
		\State $N \gets $\Call{SAMPLE}{$\mathcal{P}_{\rm in}$}
		\For{$n \gets 1$ to $N$}~~//$N$ \textit{is instance number per image}
		\State $\rm class \gets $\Call{SAMPLE}{$\mathcal{P}_{\rm ic}$}
		\State $\rm{aspect~ratio} \gets $\Call{SAMPLE}{$\mathcal{P}^{\rm aspect~ratio}_{\rm sgc},\rm class$}
		\State $\rm{scale} \gets $\Call{SAMPLE}{$\mathcal{P}^{\rm scale}_{\rm sgc},\rm class$}
		\State $\rm{location} \gets $\Call{SAMPLE}{$\mathcal{P}^{\rm location}_{\rm sgc},\rm class$}
		\State $\rm L_{obj} \gets (class, aspect ratio, scale, location) $ 
		\State $\rm class \gets $\Call{SAMPLE}{$\rm{p}_{\rm id}~\rm matrix$}
		\EndFor 
		\State \Return{$\rm L_{obj}$}
	\end{algorithmic}  
	\label{alg:A1}
\end{algorithm}

Upon determining the initial object class, SCDKG proceeds to define object attributes (including aspect ratio, scale, and location) by adhering to the spatial geometric characteristics of realistic RS objects and the initial class. The spatial geometric characteristics are modeled through three core probability functions: $\mathcal{P}_{\rm sgc}$ (Spatial Geometric Characteristics) describes instance attributes through location, scale, and aspect ratio distributions. Specifically, the location function fits center point coordinates, the scale function models object lengths, and the length-to-width ratio function determines size proportions. Meanwhile, $\mathcal{P}_{\rm ic}$ (Probability of Instance Class) governs class occurrence likelihoods, $\mathcal{P}_{\rm in}$ (Probability of Instance Number) regulates per-class instance counts, and $\rm p_{id}$ (Probability of Interdependency) quantifies class co-occurrence relationships.

Concretely, it models each class object through the spatial geometric characteristics probability density function $\mathcal{P}_{\rm sgc}$. For attributes like aspect ratio and scale, SCDKG employs a one-dimensional probability density function after analyzing the relevant geometric features of all instances in the dataset. The location attribute is modeled by decomposing the instance center point coordinates into two dimensions, $x$ and $y$, and fitting them with a two-dimensional probability density function. Utilizing $\mathcal{P}^{\rm all}_{\rm sgc}$ for all RS objects, SCDKG assigns attributes according to the class-specific $\mathcal{P}_{\rm sgc}$.

After establishing the initial object class and attributes, the selection of the subsequent object class must consider existing interdependencies among different classes. Sampling directly from $\mathcal{P}_{\rm ic}$ would overlook these relationships. For instance, the likelihood of 'ship' and 'harbor' co-occurring is higher than that of 'airplane' and 'harbor'. Thus, ensuring a rational spatial distribution is crucial to bridging the gap between generated RS images and real samples, thereby maintaining the authenticity of synthesized data. SCDKG constructs a bidirectional graph to represent these interdependencies, where each node denotes an instance class present in the RS images. A directed edge from node A to node B signifies the interdependency probability $\rm p_{ id}$ of class B objects on class A objects.

The $\rm p_{ id}$ values for different object class pairs form an interdependency probability matrix (illustrated in Fig.\ref{fig:V2} $\rm p_{ id}$ matrix). Based on the previous instance class, SCDKG samples the next instance class from this matrix and assigns attributes through the corresponding $\mathcal{P}_{\rm sgc}$. Finally, the model determines a rational instance density (i.e., 'N' in Step~1 of Fig.~\ref{fig:V2}) using the probability density function $\mathcal{P}_{\rm in}$.

\subsection{Iso-Spacing Instance Map}
\begin{figure}[t]
	\centering
	\includegraphics[width=0.99\linewidth]{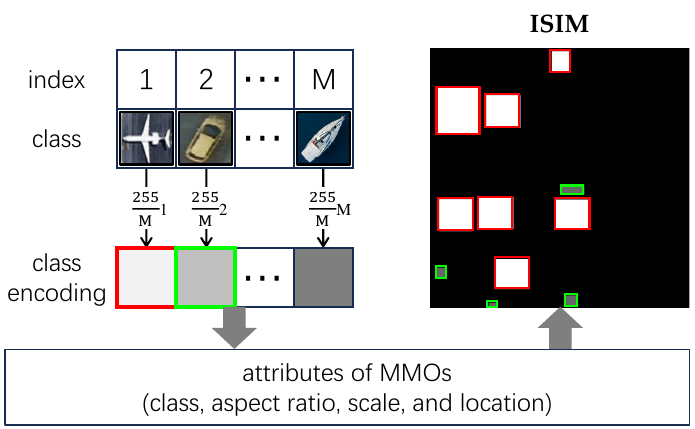}
	\caption{Illustration of the proposed ISIM encodes instances with different classes according to different grayscale values while keeping the location and scale characteristics of the corresponding regions on generated instances.}
	\label{fig:V4}
\end{figure}

Different from previous methods, the proposed MMO-IG is proposed to focus on the generation of RS images with various MMOs, which demands the control condition to represent all kinds of objects while ensuring the differences among them. Considering the generic control conditions (e.g., depth map, segment map, and edge map) lacks the ability to distinguish objects with different classes, ISIM is introduced in this paper (as shown in Fig.~\ref{fig:V4}).

Given a remote sensing (RS) dataset containing $M$ classes, the ISIM method initially assigns each class a unique numerical identifier ranging from $\left [ 0\sim {M} \right ]$. This identifier is then used in an arithmetic interval grayscale assignment strategy. The grayscale value corresponding to each class can be computed as the following equation:
\begin{equation}
	v_{gray}(m) = \left\lfloor \frac{255}{M} m \right\rfloor,
\end{equation}
where $m$ represents the unique numerical identifier for each class instance and $v_{gray}$ is the corresponding grayscale value. $ \left\lfloor \cdot \right\rfloor$ is floor function (round down to the nearest integer. With the determined class encoding for all kinds of instances in a specific dataset, MMO attributes from Step~1 in Fig.~\ref{fig:V2} are utilized to synthesize ISIM. Concretely, ISIM encodes each object based on regions corresponding to the class-specific grayscale values and spatial geometric characteristics (i.e., aspect ratio, scale, and location) on a grayscale image.

\subsection{Structured Object Distribution Instruction}
As we illustrated before, ISIM is designed to represent the spatial attributes of different RS objects and keep distinguishment among them. However, in the background region of ISIM, some RS objects would occur because of the hallucination problem of DGMs, which leads to a discrepancy between the instance-level labels (ISIM) and the generated RS image and further negatively impacts downstream tasks. 

\begin{figure}[t]
	\centering
	\includegraphics[width=0.99\linewidth]{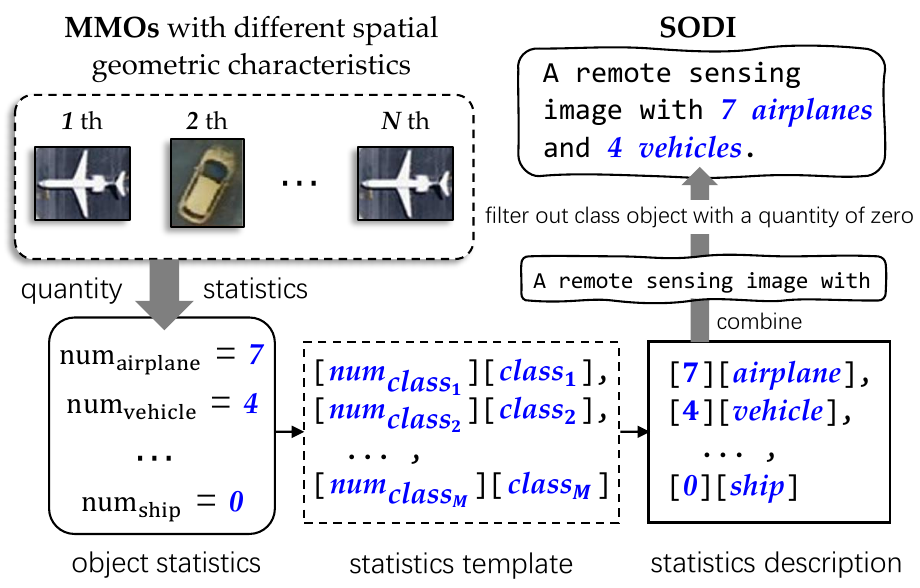}
	\caption{Illustration of the proposed SODI generation process. It consists of the combination of a structured scene head description (`` A remote sensing image with'') and a statistics description of RS objects.}
	\label{fig:V5}
\end{figure}

Therefore, SODI is introduced to ensure the alignment between ISIM and the generated RS image. As shown in Fig.~\ref{fig:V5}, with the determined MMOs with different spatial geometric characteristics from Step~1 in Fig.~\ref{fig:V2}, the SODI generation process first counts the number of objects of different classes and fills the information into the statistics template for obtaining a statistics description about MMOs that from SCDKG. Then, the class object with a quantity of zero in the statistics description is filtered out and the filtered description is combined with a structured scene head description (`` A remote sensing image with'') for generating the SODI. The scene head description assists our model in regulating image content to align with the remote sensing style. Meanwhile, the filtered statistics description ensures that MMO-IG maintains strict alignment between MMOs in ISIM and the generated remote sensing images. This guidance, from a global perspective, facilitates an accurate and controllable image generation process.

\subsection{Decoding ISIM to Generate RS Image}
The popular DGMs, diffusion models~\cite{sohl2015deep,ho2020denoising,rombach2022high,zhang2023adding}, reconstruct images via a denoising process on noisy counterparts. Considering the diffusion models' advantages of stable training, high-quality sample generation, tractable likelihood estimation, versatility, and robustness to hyperparameter settings compared with VAE~\cite{DBLP:journals/corr/KingmaW13} and GAN~\cite{DBLP:conf/nips/GoodfellowPMXWOCB14} models, MMO-IG follows the stable diffusion~\cite{rombach2022high} structure to construct the corresponding generative network. As shown in Step~3 of Fig.~\ref{fig:V2}, the generative part consists of two couples of res encoder and res decoder mainly. 

In designing our residual encoder (res encoder) and residual decoder (res decoder), we drew inspiration from the architecture of the stable diffusion model. Specifically, the res encoder and res decoder each contain 12 blocks, with the entire model comprising 25 blocks, including the middle block. Among these 25 blocks, 8 blocks are convolutional layers responsible for downsampling or upsampling, while the remaining 17 blocks are the main functional blocks. Each main block consists of 4 residual network layers and 2 Vision Transformers (ViTs). Notably, each ViT incorporates multiple cross-attention and self-attention mechanisms, which further enhance the model's expressive power and flexibility.

The encoder and decoder are composed of 12 stable diffusion layers respectively and they are connected by a middle layer that of 1 stable diffusion layer. The frozen generative blocks are initialized through the pre-trained weights from stable diffusion~\cite{rombach2022high}. The structures of trainable generative blocks are copied from the right ones and the two block weights are initialized by coping the right block and zero, respectively. 

In the generation process, the generative network takes ISIM, SODI, and $z^T$ as input to achieve the RS images, where ISIM and SODI are embedding into feature vectors via a lightweight network $\varepsilon$ and CLIP~\cite{DBLP:conf/icml/RadfordKHRGASAM21} respectively, where $\varepsilon$ consists of four convolution layers with 4$\times$4 kernels and 2$\times$2 strides $\rm Conv^{16}_{4,2}, Conv^{32}_{4,2}, Conv^{64}_{4,2}, and Conv^{128}_{4,2}$. $z^T$ is the feature vector of noisy image that can be efficiently obtained by conducting the well-trained $\varepsilon$ on the noisy image. After T-1 times iterative denoising through res blocks, MMO-IG decoding each region in ISIM into the corresponding object with the decoded image feature vector $z$, and the final RS image reconstructed by conducting $D$ (that proposed in ~\cite{DBLP:conf/cvpr/EsserRO21}) on $z$.

Given the clean input image $z_0$, diffusion models gradually corrupt the image through $t$ successive noising steps, producing the noisy latent $z_t$ where $t \in \{1,...,T\}$ denotes the timestep index. Conditioned on the timestep $t$,  structured object distribution instruction  (SODI) $c_t$, and iso-spacing instance map (ISIM) $c_f$, the model learns a denoising network $\epsilon_\theta$ to estimate the injected noise in $z_t$. This process is formalized as:
\begin{equation}
\mathcal{L} = \mathbb{E}_{z_0, t, c_t, c_f, \epsilon \sim \mathcal{N}(0,\mathbf{I})} \left[ \left\| \epsilon - \epsilon_\theta(z_t, t, c_t, c_f) \right\|_2^2 \right]
\end{equation}
where $\mathcal{L}$ denotes the loss function for training the diffusion model.

\section{Experiments}
\label{Experiments}
In this section, we detail the dataset configuration used for our experiments, present the experimental results of our MMO-IG model, and conduct an ablation study. Additionally, we design downstream experiments to demonstrate the effectiveness of the generated remote sensing multi-object detection data using this method, achieving results comparable to those obtained with real remote sensing images.

\subsection{Datasets and Experimental Setup}
\textbf{Datasets.}
We utilized the DIOR and DIOR-R \cite{cheng2022anchor} dataset to train and evaluate our model. This dataset comprises 23,463 high-quality remote sensing images and 190,288 meticulously annotated object instances, resulting in a total of 192,472 axis-aligned object annotations. Each image is 800×800 pixels, with spatial resolutions ranging from 0.5 to 30 meters. The dataset is divided into a training and validation set (11,725 images) and a test set (11,738 images). DIOR serves as a comprehensive benchmark for object detection in optical remote sensing images, encompassing 20 object classes, including \textbf{A}irplane, \textbf{A}irport, \textbf{B}aseball, \textbf{B}asketball, \textbf{B}ridge, \textbf{C}himney, \textbf{D}am, \textbf{E}xpressway service area, \textbf{E}xpressway toll station, \textbf{G}olf field, \textbf{G}round track, \textbf{H}arbor, \textbf{O}verpass, \textbf{S}hip, \textbf{S}tadium, \textbf{S}torage tank, \textbf{T}ennis court, \textbf{T}rain station, \textbf{V}ehicle, \textbf{W}indmill.
\begin{figure*}[t]
	\centering
	\includegraphics[width=0.99\linewidth]{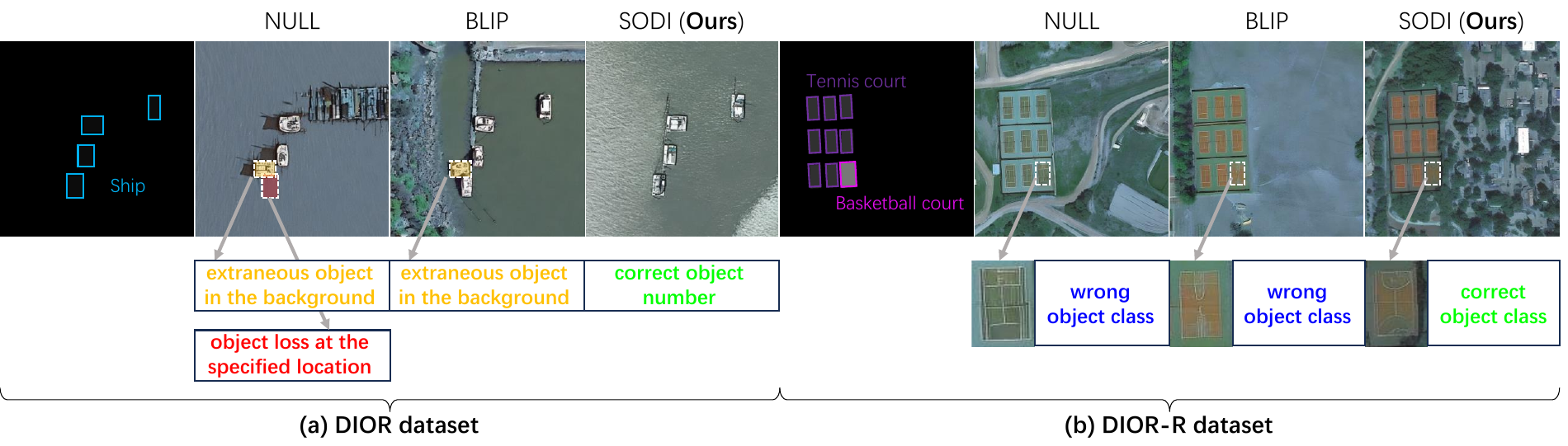}
	\caption{Visualization comparison of the generated image by our MMO-IG with different text conditions on the DIOR dataset. Considering the grayscale values of some objects are too low and difficult to discern, we enhance their visibility by marking the object regions with colored bounding boxes in the ISIM.}
	\label{fig:V6}
\end{figure*}

\begin{table}[]
	\renewcommand{\arraystretch}{1.2}
	\setlength{\tabcolsep}{1.3mm}
	\caption{Results on MMO-IG with different text conditions. `NULL' and `BLIP' denote the text condition is null and an image description generated by BLIP~\cite{li2022blip}. $\rm Acc_{c}$ and $\rm Acc_{n}$ are the accuracy of the object class and number that are evaluated on the generated RS images.}
	\centering
	\begin{tabular}{clcccc}
		\toprule
		\rowcolor{grayyy} dataset& text condition & $\rm Acc_{c}\uparrow$ & $\rm Acc_{n}\uparrow$ & $\rm FID\downarrow$ & CAS$\uparrow$ \\ \midrule
		\multirow{3}{*}{DIOR}   & NULL           &     92.8     &    93.9    &  128.4   &   27.1       \\
		& BLIP           & 92.3    & 94.5  &   75.6  &    38.4      \\ 
		& SODI (\textbf{Ours})     & \textbf{97.9}(\textcolor{greenn}{\textbf{5.6}})    & \textbf{98.7}(\textcolor{greenn}{\textbf{4.2}})  &  \textbf{65.6}(\textcolor{greenn}{\textbf{10.0}})   &    \textbf{45.8}(\textcolor{greenn}{\textbf{7.4}})      \\ \midrule
		\multirow{3}{*}{DIOR-R} & NULL           &     91.7     &    93.7    &  134.6   &   28.3       \\ 
		& BLIP           &     93.4     &   93.1    &  69.8   &     36.9     \\ 
		& SODI (\textbf{Ours})     &    \textbf{98.2}(\textcolor{greenn}{\textbf{4.8}})      &   \textbf{97.2}(\textcolor{greenn}{\textbf{4.1}})    &  \textbf{64.8}(\textcolor{greenn}{\textbf{5.0}})   &   \textbf{47.2}(\textcolor{greenn}{\textbf{10.3}})      \\ \bottomrule
	\end{tabular}
	\label{tab:T1}
\end{table}

\textbf{Experimental Setup.}
Our model is trained using the Adam optimizer with a learning rate of 1e-5. Training is conducted on four NVIDIA Tesla A100 GPUs, each with 80 GB of memory, and takes approximately two days to complete. During the sampling phase, six samples are generated, with a control parameter (scale) set to 2.5 to enhance sampling quality and refine the model's outputs. This configuration allows for efficient utilization of computational resources, achieving high performance in both the training and sampling stages.
\subsection{Ablation Study}
\textbf{Effectiveness of SODI.} Considering the hallucination problem in deep generative models (DGMs) can cause remote sensing (RS) objects to be inaccurately represented in the background, resulting in inconsistencies between instance-level labels (ISIM) and the generated RS images (as shown in Fig.~\ref{fig:V6}). When low-quality ISIM is used as a control condition for generating images, it leads to a mismatch between the generated images and the provided SODI. This mismatch negatively affects downstream tasks, such as object detection, since the labels for the generated detection data are derived from the SODI. If the generated images do not align with the SODI, the resulting data becomes unreliable, which can significantly impair the model's performance during training. This discrepancy adversely affects downstream tasks. Therefore,  SODI is implemented to guide the image generation process by incorporating a global perspective on image style and instance characteristics, thereby ensuring precision and control. To verify the effectiveness of SODI in ensuring the alignment between ISIM and the generated RS images, we show the testing quantitative metrics of our model with different text conditions in Table~\ref{tab:T1}, where `NULL' and `BLIP' denote the text condition is null and an image description generated by BLIP~\cite{li2022blip}. It can be found that MMO-IG with `NULL' text condition makes it hard to control the accuracy of object class and number on the generated RS images, where the $\rm Acc_c$ and $\rm Acc_n$ are 92.8 and 93.9 respectively. It means that ISIM, the instance-level label, enjoys low-quality correspondence with the generated RS images, further influencing the following downstream task. $\rm Acc_c $ evaluates category alignment between generated instances and SODI-defined classes, while $\rm Acc_n$ measures numerical consistency in instance counts relative to SODI specifications.

As for the `BLIP' text condition, it can describe image content in detail, which provides a more controlled ability to achieve a realistic image style. It can be found from Table~\ref{tab:T1}, that the `BLIP' text condition brings more than 50 and 8 improvements in FID and CAS respectively, which means there is a significant gain in the similarity between generated images and real images. However, in generating object detection data, the `BLIP' text condition often presents challenges. Specifically, it typically emphasizes the overall features of the image, such as color, background, or scene, while paying insufficient attention to the categories and quantities of instances within the image. For instance, BLIP might generate a descriptive text like "A beautiful aerial view of a city with roads and buildings," but it rarely accurately identifies specific object categories (such as "cars," "tennis courts," etc.) or the number of objects in the image. Consequently, the text generated by BLIP may lack the precision necessary, especially when such information is crucial for training object detection models. It leads to the problem of low-quality correspondence between ISIM and the generated RS image is still under-alleviated. The above phenomenon can be verified by the results in Table~\ref{tab:T1}, compared with `NULL' text condition, the $\rm Acc_c$ and $\rm Acc_n$ of the model with `BLIP' text condition not achieve significant improvements, and is even slightly degraded.
\begin{figure}[t]
	\centering
	\includegraphics[width=0.99\linewidth]{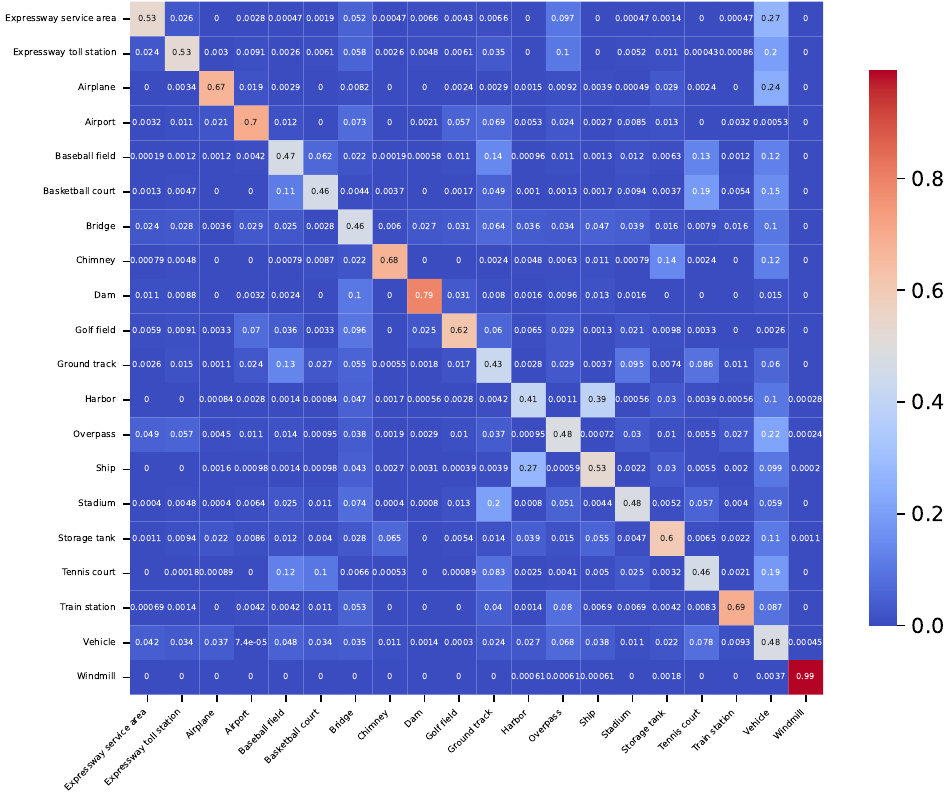}
	\caption{Visualization of the $\rm p_{id}$ matrix of SCDKG on DIOR dataset. There are 20 classes in the dataset and the value of each element represents the interdependencies between the corresponding two classes.}
	\label{fig:V8}
\end{figure}

Different from the `BLIP' text condition, SODI  begins with "a remote sensing image with," followed by a detailed description based on the actual instance categories and quantities present in the image. For example, if there are 37 ships and 3 harbors in the image, the generated prompt would be a remote sensing image with 37 ships, and 3 harbors, if there are 4 bridges and 3 ground track fields in the image, the generated prompt would be: a remote sensing image with 4 bridges, 3 ground track fields. It ensures that the generated text not only accurately describes the overall features of the image but also provides a detailed account of the specific object types and quantities contained within the image, thereby offering more precise training data for the model. Models trained using this method can effectively eliminate errors related to instance quantity and category during the generation process. During the image generation process, the model no longer makes errors due to inaccurate descriptions of object types or quantities. This method resolves instance recognition errors in the generation of RS images with MMOs, ensuring that the strict correspondences between ISIM and generated images better meet the practical requirements of downstream object detection tasks. These advantages help our MMO-IG achieve 97.9 $\rm Acc_c$ and 98.7 $\rm Acc_n$ on the DIOR dataset and 98.2 $\rm Acc_c$ and 97.2 $\rm Acc_n$ on the DIOR-R dataset, which enhances the correspondence between ISIM and the generated RS image significantly. Moreover, our model surpasses the `BLIP' text condition a lot in FID and CAS, which means the SODI's ability to improve the reality of the generated image. The superiority also has been verified by the qualified generated RS image samples in Fig.~\ref{fig:V6}. The above results show that SODI can improve the generation effect, making the image content real, and the alignment between ISIM and generated RS images.
\begin{figure}[t]
	\centering
	\includegraphics[width=0.99\linewidth]{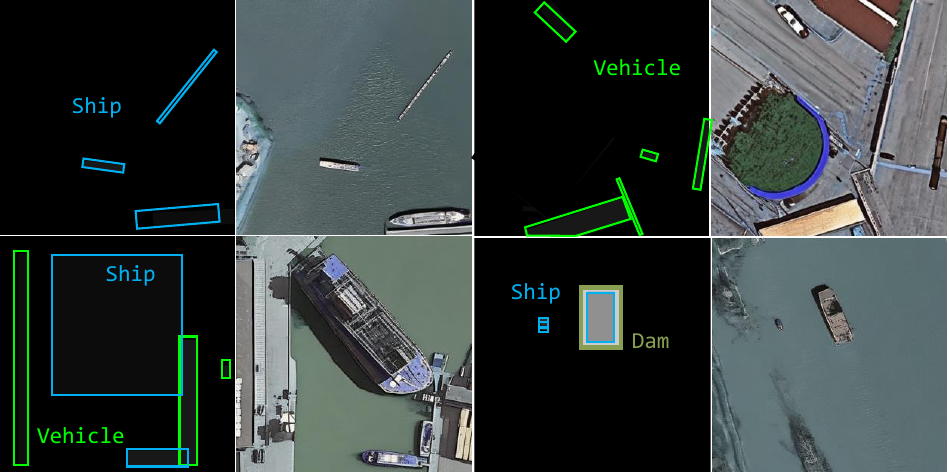}
	\caption{Visualization of some samples with irrational aspect ratio, scale, and locations without the guidance of SCDKG. Considering the grayscale values of some objects are too low and difficult to discern, we enhance their visibility by marking the object regions with colored bounding boxes in the ISIM.}
	\label{fig:V7-1}
\end{figure}

\begin{table}[]
	\renewcommand{\arraystretch}{1.2}
	\setlength{\tabcolsep}{.95mm}
	\caption{Results of models with different SCDKG settings on DIOR dataset. $\rm FID_{zs}$ and $\rm CAS_{zs}$ denote the zero-shot FID and CAS metrics.}
	\centering
	\begin{tabular}{lcccccccc}
		\toprule
		\multirow{2}{*}{\#} & \multirow{2}{*}{\begin{tabular}[c]{@{}c@{}}aspect\\ ratio\end{tabular}} & \multirow{2}{*}{scale} & \multirow{2}{*}{location} & \multirow{2}{*}{\begin{tabular}[c]{@{}c@{}}$\rm p_{id}$\\ matrix\end{tabular} } & \multicolumn{2}{c}{DIOR}      & \multicolumn{2}{c}{DIOR-R}    \\ \cline{6-9} 
		&                               &                        &                           &                      & \multicolumn{1}{c}{$\rm FID_{zs}$$\downarrow$} & $\rm CAS_{zs}$$\uparrow$ & \multicolumn{1}{c}{$\rm FID_{zs}$$\downarrow$} & $\rm CAS_{zs}$$\uparrow$ \\ \midrule
		1&                               &                        &                           &                      & \multicolumn{1}{c}{92.3}    &   28.2  & \multicolumn{1}{c}{94.2}    &  27.7   \\ 
		2&                               &                        &                           &    \ding{52}        & \multicolumn{1}{c}{88.5}    &  29.3   & \multicolumn{1}{c}{87.4}    &   28.7  \\ 
		3&      \ding{52}               &                        &                           &     \ding{52}        & \multicolumn{1}{c}{82.7}    &  30.7   & \multicolumn{1}{c}{83.9}    &   29.8  \\ 
		4&                               &     \ding{52}        &                           &     \ding{52}        & \multicolumn{1}{c}{74.5}    &  32.4   & \multicolumn{1}{c}{73.5}    &   31.9  \\ 
		5&                               &                        &      \ding{52}         &     \ding{52}        & \multicolumn{1}{c}{79.7}    &  34.9   & \multicolumn{1}{c}{78.2}    &  35.3   \\ 
		6&     \ding{52}               &       \ding{52}       &                           &     \ding{52}        & \multicolumn{1}{c}{70.2}    &   38.2  & \multicolumn{1}{c}{71.2}    &   37.9  \\ 
		7&                               &    \ding{52}          &     \ding{52}          &     \ding{52}        & \multicolumn{1}{c}{78.9}    &   40.1  & \multicolumn{1}{c}{74.5}    &   42.3  \\ 
		8&       \ding{52}             &                        &      \ding{52}           &     \ding{52}        & \multicolumn{1}{c}{68.7}    &   44.6  & \multicolumn{1}{c}{67.2}    &   46.2  \\ 
		9&      \ding{52}              &   \ding{52}          &       \ding{52}          &     \ding{52}        & \multicolumn{1}{c}{\underline{\textbf{65.6}}}    &   \underline{\textbf{45.8}}  & \multicolumn{1}{c}{\underline{\textbf{64.8}}}    &   \underline{\textbf{47.2}}  \\ \bottomrule
	\end{tabular}
	\label{tab:T2}
\end{table}

\textbf{Effectiveness of SCDKG.} To address complex interdependencies among objects of different classes and their diverse spatial geometric characteristics (as illustrated in Fig.~\ref{fig:V3}). SCDKG is proposed to ensure a rational and realistic content distribution for the generated RS images.

SCDKG consists of $\rm p_{id}$ matrix that models interdependencies among all objects of different classes (the $\rm p_{id}$ matrix of DIOR dataset is shown in Fig.~\ref{fig:V8}) and $\mathcal{P}_{\rm sgc}$ that formulates diverse spatial geometric characteristics of each object class individually. Notably, $\mathcal{P}_{\rm sgc}$ models the aspect ratio, scale, and location characteristics of RS objects separately. It effectively avoids influence brought by irrational characteristics (e.g., aspect ratio and positional distribution that do not conform to real-world object spatial geometric characteristics visualized in Fig.~\ref{fig:V7-1}). To verify the effectiveness of SCDKG, we test the performance enhancement brought by each factor in SCDKG and visualize the results in Table~\ref{tab:T2}. For $\mathcal{P}_{\rm sgc}$, It can be found that all of the factors can bring performance gains for the generated RS images. Specifically, the modeling of the complex interdependencies among different RS objects makes a more rational object distribution, which helps our model achieve 3.8 and 1.1 improvements in $\rm FID_{zs}$ and CAS on DIOR dataset and 6.8 and 1.0 gains in $\rm FID_{zs}$ and CAS on DIOR-R dataset. Notably, $\rm FID_{zs}$ and $\rm CAS_{zs}$ denote the zero-shot FID and zero-shot CAS, which is introduced to evaluate the quality of the generated images that not in the test set. The improvements brought by $\rm p_{id}$ matrix demonstrate the effectiveness of the interdependencies modeling. 
\begin{figure*}[t]
	\centering
	\includegraphics[width=0.99\linewidth]{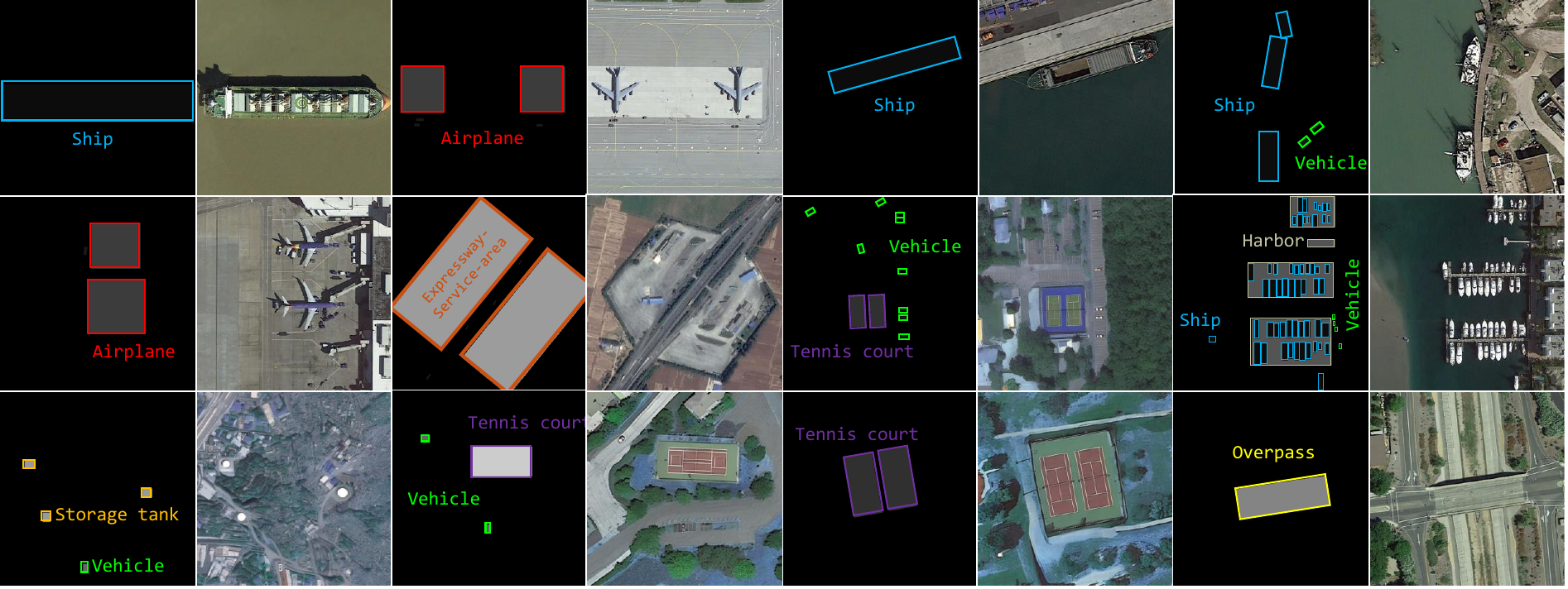}
	\caption{Visualization of some qualified generated RS images by our MMO-IG. Considering the grayscale values of some objects are too low and difficult to discern, we enhance their visibility by marking the object regions with colored bounding boxes in the ISIM.}
	\label{fig:V7}
\end{figure*}

Besides, the rational characteristics of the object aspect ratio, scale, and location bring 8.8, 5.8, and 14.0 improvements in $\rm FID_{zs}$ and 1.4, 3.1, and 5.6 gains in $\rm CAS_{zs}$. This is mainly because the synthesized images enjoy more rational and realistic object spatial characteristics than real RS images, which makes it easier for our model to learn the feature distribution and bring significant performance improvements. Meanwhile, the combination of the three characteristics also can bring further improvements in both $\rm FID_{zs}$ and $\rm CAS_{zs}$. Specifically, the model equipped with SCDKG gains 26.7 and 17.6 in $\rm FID_{zs}$ and $\rm CAS_{zs}$ on DIOR dataset and 29.4 and 19.5 in $\rm FID_{zs}$ and $\rm CAS_{zs}$ on DIOR-R dataset. Significant performance improvements verify the effectiveness of the proposed SCDKG in the generation task of RS images containing MMOs. Some qualified samples are shown in Fig.~\ref{fig:V7}, it can be found that SCDKG enables ensuring rational spatial geometric characteristics for objects with different classes to enhance the reality of generated images.
\begin{table}[]
 	\renewcommand{\arraystretch}{1.2}
 	\setlength{\tabcolsep}{2.6mm}
 	\caption{Comparisons with existing layout-to-image methods on DIOR and DIOR-R datasets.}
 	\centering
 	\begin{tabular}{clcc}
 		\toprule
 		\rowcolor{grayyy}dataset                 & method           & $\rm FID\downarrow$   & CAS$\uparrow$   \\ \midrule
 		\multirow{5}{*}{DIOR}   & LostGAN~\cite{sun2019image}          & 57.10 & 46.02 \\ 
 		& Layout Diffusion~\cite{zheng2023layoutdiffusion} & 45.31 & 56.98 \\ 
 		& ReCo~\cite{yang2023reco}             & 42.56 & 55.42 \\ 
 		& GLIGEN~\cite{li2023gligen}           & 41.31 & 63.50 \\ 
 		& MMO-IG (\textbf{Ours})    & \textbf{34.48}(\textcolor{greenn}{\textbf{6.83}}) & \textbf{78.64}(\textcolor{greenn}{\textbf{15.14}}) \\ \midrule
 		\multirow{2}{*}{DIOR-R} & GLIGEN~\cite{li2023gligen}           & 48.43 & 58.89 \\ 
 		& MMO-IG (\textbf{Ours})    & \textbf{35.07}(\textcolor{greenn}{\textbf{13.36}}) & \textbf{76.13}(\textcolor{greenn}{\textbf{17.24}}) \\ \bottomrule
 	\end{tabular}
 	\label{tab:T3}
\end{table}

\subsection{Comparisons with Existing DGMs} 
Different from existing layout-to-image DGMs, our MMO-IG enables us to focus on the complex interdependencies among different RS objects, which helps ensure the rational distribution of the generated RS images. Meanwhile, the SODI can suppress the appearance of objects in the background. The above advantages facilitate MMO-IG's superior performance in the generation process. In this section, we compare our MMO-IG with existing layout-to-image DGMs to show the superiority of our method. As shown in Table~\ref{tab:T3}, for horizontal labeled samples (DIOR data), our method achieves 6.83 and 15.14 improvements in FID and CAS evaluation metrics respectively. It is mainly because of the more direct encoding strategy (ISIM) for RS objects with different classes and the more effective constraints on the object quantity and class brought by SODI. The former helps our model decode ISIM regions into instances following the simple correspondence between the classes and grayscale values, which is more intuitive than the way to represent instance class through text description. The latter ensures a strict correspondence of RS objects between ISIM and generated RS images. The above advantages either facilitate MMO-IG to surpass GLIGEN~\cite{li2023gligen} 13.36 and 17.24 in the aspect of FID and CAS respectively on rotated labeled data (DIOR-R). The results demonstrate the superiority of our MMO-IG and the effectiveness of the introduced ISIM and SODI on the generation task of RS images that contain MMOs.
\begin{figure*}[t]
	\centering
	\includegraphics[width=0.95\linewidth]{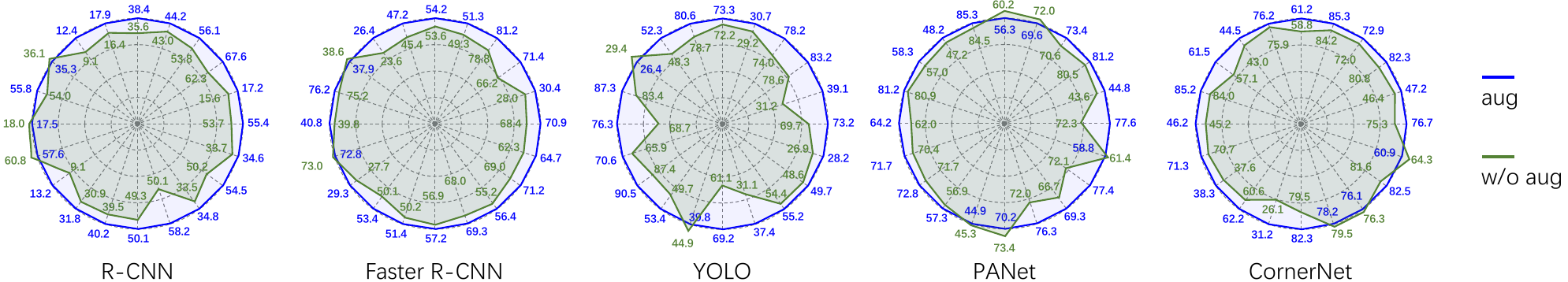}
	\caption{Comparison of different detectors' accuracy across each category on the DIOR dataset under the setting of data augmentation with 20k generation images (aug) and without augmentation (w/o aug). There are 20 classes of objects (i.e. Airplane, Airport, Baseball, Basketball, Bridge, Chimney, Dam, Expressway service area, Expressway toll station, Golf field Ground track, Harbor, Overpass, Ship, Stadium, Storage tank, Tennis court, Train station, Vehicle, Windmill) in DIOR dataset. The performance metrics for these classes are arranged clockwise from the 12 o'clock position in the diagram.}
	\label{fig:V9}
\end{figure*}

\begin{table*}[]
	\renewcommand{\arraystretch}{1.2}
	\setlength{\tabcolsep}{1.mm}
	\caption{Results of different detectors' accuracy across each category on the DIOR dataset under the setting of data augmentation with 20k generation images (aug) and without augmentation (w/o aug). \textbf{E}xpressway s-a and \textbf{E}xpressway t-s denote \textbf{E}xpressway service area and \textbf{E}xpressway toll station.}
	\centering
	\begin{tabular}{lccccccccccccc}
		\toprule
		\rowcolor{grayyy}methods                       & aug & \textbf{A}irplane     & \textbf{A}irport & \textbf{B}aseball & \textbf{B}asketball & \textbf{B}ridge  & \textbf{C}himney      & \textbf{D}am          & \textbf{E}xpressway s-a & \textbf{E}xpressway t-s & \textbf{G}olf field & \\ \midrule
		\multirow{2}{*}{R-CNN~\cite{girshick2014rich}}        &   \ding{56}   & 35.6 & 43.0 & 53.8 & 62.3 & 15.6 & 53.7 & 33.7 & 50.2 & 33.5 & 50.1 &{}    \\  
		&  \ding{52}   &  \textcolor{greenn}{\textbf{38.4}} & \textcolor{greenn}{\textbf{44.2}} & \textcolor{greenn}{\textbf{56.1}} & \textcolor{greenn}{\textbf{67.6}} & \textcolor{greenn}{\textbf{17.2}} & \textcolor{greenn}{\textbf{55.4}} & \textcolor{greenn}{\textbf{34.6}} & \textcolor{greenn}{\textbf{54.5}} & \textcolor{greenn}{\textbf{34.8}} & \textcolor{greenn}{\textbf{58.2}} &{}\\ 
		\multirow{2}{*}{Faster R-CNN~\cite{ren2016faster}} &  \ding{56}   & 53.6 & 49.3 & 78.8 & 66.2 & 28.0 & 68.4 & 62.3 & 69.0 & 55.2 & 68.0 &{} \\  
		&  \ding{52}   & \textcolor{greenn}{\textbf{54.2}} & \textcolor{greenn}{\textbf{51.3}} & \textcolor{greenn}{\textbf{81.2}} & \textcolor{greenn}{\textbf{71.4}} & \textcolor{greenn}{\textbf{30.4}} & \textcolor{greenn}{\textbf{70.9}} & \textcolor{greenn}{\textbf{64.7}} & \textcolor{greenn}{\textbf{71.2}} & \textcolor{greenn}{\textbf{56.4}} & \textcolor{greenn}{\textbf{69.3}} &{}\\ 
		\multirow{2}{*}{YOLO~\cite{redmon2018yolov3}}         &   \ding{56}  & 72.2 & 29.2 & 74.0 & 78.6 & 31.2 & 69.7 & 26.9 & 48.6 & 54.4 & 31.1  &{}\\  
		&  \ding{52}   & \textcolor{greenn}{\textbf{73.3}} & \textcolor{greenn}{\textbf{30.7}} & \textcolor{greenn}{\textbf{78.2}} & \textcolor{greenn}{\textbf{83.2}} & \textcolor{greenn}{\textbf{39.1}} & \textcolor{greenn}{\textbf{73.2}} & \textcolor{greenn}{\textbf{28.2}} & \textcolor{greenn}{\textbf{49.7}} & \textcolor{greenn}{\textbf{55.2}} & \textcolor{greenn}{\textbf{37.4}} &{}\\ 
		\multirow{2}{*}{PANet~\cite{liu2018path}}        &  \ding{56}   & 60.2 & 72.0 & 70.6 & 80.5 & 43.6 & 72.3 & 61.4 & 72.1 & 66.7 & 72.0 &{}\\  
		&   \ding{52}  & \textcolor{redd}{\textbf{56.3}} & \textcolor{redd}{\textbf{69.6}} & \textcolor{greenn}{\textbf{73.4}} & \textcolor{greenn}{\textbf{81.2}} & \textcolor{greenn}{\textbf{44.8}} & \textcolor{greenn}{\textbf{77.6}} & \textcolor{redd}{\textbf{58.8}} & \textcolor{greenn}{\textbf{77.4}} & \textcolor{greenn}{\textbf{69.3}} & \textcolor{greenn}{\textbf{76.3}} &{}\\ 
		\multirow{2}{*}{CornerNet~\cite{law2018cornernet}}    &   \ding{56}  & 58.8 & 84.2 & 72.0 & 80.8 & 46.4 & 75.3 & 64.3 & 81.6 & 76.3 & 79.5 &{}\\  
		&   \ding{52}  &\textcolor{greenn}{\textbf{61.2}} & \textcolor{greenn}{\textbf{85.3}} & \textcolor{greenn}{\textbf{72.9}} & \textcolor{greenn}{\textbf{82.3}} & \textcolor{greenn}{\textbf{47.2}} & \textcolor{greenn}{\textbf{76.7}} & \textcolor{redd}{\textbf{60.9}} & \textcolor{greenn}{\textbf{82.5}} & \textcolor{redd}{\textbf{76.1}} & \textcolor{redd}{\textbf{78.2}} &{}\\ \midrule
		
		\rowcolor{grayyy}methods                       & aug & \textbf{G}round track & \textbf{H}arbor  & \textbf{O}verpass & \textbf{S}hip       & \textbf{S}tadium & \textbf{S}torage tank & \textbf{T}ennis court & \textbf{T}rain station   & \textbf{V}ehicle                 & \textbf{W}indmill & \textbf{A}vg  \\ \midrule
		\multirow{2}{*}{R-CNN~\cite{girshick2014rich}}        &   \ding{56}  & 49.3 & 39.5 & 30.9 & 9.1 & 60.8 & 18.0 & 54.0 & 36.1 & 9.1 & 16.4 &37.1\\  
		&   \ding{52}  & \textcolor{greenn}{\textbf{50.1}} & \textcolor{greenn}{\textbf{40.2}} & \textcolor{greenn}{\textbf{31.8}} & \textcolor{greenn}{\textbf{13.2}} & \textcolor{redd}{\textbf{57.6}} & \textcolor{redd}{\textbf{17.5}} & \textcolor{greenn}{\textbf{55.8}} & \textcolor{redd}{\textbf{35.3}} & \textcolor{greenn}{\textbf{12.4}} & \textcolor{greenn}{\textbf{17.9}}&\textcolor{greenn}{\textbf{37.7}} \\ 
		\multirow{2}{*}{Faster R-CNN~\cite{ren2016faster}} &   \ding{56}  & 56.9 & 50.2 & 50.1 & 27.7 & 73.0 & 39.8 & 75.2 & 38.6 & 23.6 & 45.4&53.4\\  
		&   \ding{52}  & \textcolor{greenn}{\textbf{57.2}} & \textcolor{greenn}{\textbf{51.4}} & \textcolor{greenn}{\textbf{53.4}} & \textcolor{greenn}{\textbf{29.3}} & \textcolor{redd}{\textbf{72.8}} & \textcolor{greenn}{\textbf{40.8}} & \textcolor{greenn}{\textbf{76.2}} & \textcolor{redd}{\textbf{37.9}} & \textcolor{greenn}{\textbf{26.4}} & \textcolor{greenn}{\textbf{47.2}}&\textcolor{greenn}{\textbf{54.0}} \\ 
		\multirow{2}{*}{YOLO~\cite{redmon2018yolov3}}         &   \ding{56}  & 61.1 & 44.9 & 49.7 & 87.4 & 65.9 & 68.7 & 83.4 & 29.4 & 48.3 & 78.7 &55.3\\  
		&  \ding{52}   & \textcolor{greenn}{\textbf{69.2}} & \textcolor{redd}{\textbf{39.8}} & \textcolor{greenn}{\textbf{53.4}} & \textcolor{greenn}{\textbf{90.5}} & \textcolor{greenn}{\textbf{70.6}} & \textcolor{greenn}{\textbf{76.3}} & \textcolor{greenn}{\textbf{87.3}} & \textcolor{redd}{\textbf{26.4}} & \textcolor{greenn}{\textbf{52.3}} & \textcolor{greenn}{\textbf{80.6}} &\textcolor{greenn}{\textbf{55.7}}\\ 
		\multirow{2}{*}{PANet~\cite{liu2018path}}        &   \ding{56}  & 73.4 & 45.3 & 56.9 & 71.7 & 70.4 & 62.0 & 80.9 & 57.0 & 47.2 & 84.5 &62.9\\  
		&   \ding{52}  & \textcolor{redd}{\textbf{70.2}} & \textcolor{redd}{\textbf{44.9}} & \textcolor{greenn}{\textbf{57.3}} & \textcolor{greenn}{\textbf{72.8}} & \textcolor{greenn}{\textbf{71.7}} & \textcolor{greenn}{\textbf{64.2}} & \textcolor{greenn}{\textbf{81.2}} & \textcolor{greenn}{\textbf{58.3}} & \textcolor{greenn}{\textbf{48.2}} & \textcolor{greenn}{\textbf{85.3}} &\textcolor{greenn}{\textbf{63.2}}\\ 
		\multirow{2}{*}{CornerNet~\cite{law2018cornernet}}    &   \ding{56}  & 79.5 & 26.1 & 60.6 & 37.6 & 70.7 & 45.2 & 84.0 & 57.1 & 43.0 & 75.9&62.4 \\  
		&   \ding{52}  & \textcolor{greenn}{\textbf{82.3}} & \textcolor{greenn}{\textbf{31.2}} & \textcolor{greenn}{\textbf{62.2}} & \textcolor{greenn}{\textbf{38.3}} & \textcolor{greenn}{\textbf{71.3}} & \textcolor{greenn}{\textbf{46.2}} & \textcolor{greenn}{\textbf{85.2}} & \textcolor{greenn}{\textbf{61.5}} & \textcolor{greenn}{\textbf{44.5}} & \textcolor{greenn}{\textbf{76.2}} &\textcolor{greenn}{\textbf{62.5}}\\ \bottomrule
	\end{tabular}
	\label{tab:T4}
\end{table*}

\subsection{Downstream Task} 
As we mentioned before, RSIOD~\cite{li2023instance,zhang2023efficient,gao2023global,li2023large} is an important task in the research of remote sensing. However, the high cost of acquiring satellite RS images and the labor-intensive nature of image annotation limit the availability of data for this research area, making it challenging to adequately train algorithms. Based on the above consideration, MMO-IG is proposed to alleviate the data limitation problem. 

In this section, we demonstrate the effectiveness of the generated remote sensing (RS) images using our MMO-IG model. We generated 20,000 images, which were integrated with the original DIOR dataset to train various models, including R-CNN~\cite{girshick2014rich}, Faster R-CNN~\cite{ren2016faster}, YOLO~\cite{redmon2018yolov3}, PANet~\cite{liu2018path}, and CornerNet~\cite{law2018cornernet}. These models were then evaluated on the DIOR test set. The results, presented in Table \ref{tab:T4} and Fig.~\ref{fig:V9} (for better observing the gains of all kinds of objects), compare models trained solely on the DIOR training set with those trained on the augmented dataset. The first row for each method indicates the accuracy of models trained exclusively on the DIOR dataset, while the second row shows the results with the combined generated images and DIOR dataset.

The findings reveal that the augmented data from our MMO-IG model significantly enhances performance for the vast majority of objects compared to models trained only on the DIOR dataset. For instance, R-CNN, Faster R-CNN, YOLO, PANet, and CornerNet exhibit performance gains of up to 8.1, 5.2, 8.1, 5.3, and 5.1 percentage points in detection accuracy, respectively. This underscores the effectiveness of the generated RS images in improving the performance of remote sensing image object detection (RSIOD) models.

Although there is a decline in performance for a small number of object classes (averaging 15\% per method), the RS data generated by MMO-IG still provides substantial benefits for these detectors. Additionally, we visualize the detection performance in Fig.~\ref{fig:V6} to better illustrate the improvements from synthesized data across different object classes. These results demonstrate the capability of MMO-IG to alleviate data limitations in RSIOD.

\section{Conclusion}
\label{Conclusion}
In this paper, we present a generative framework, MMO-IG, designed to synthesize remote sensing (RS) images with multi-modal objects (MMOs) and provide corresponding instance-level labels for Remote Sensing Image Object Detection (RSIOD). We introduce ISIM and SODI as control conditions to encode MMO and image content, and propose SCDKG to model interdependencies between object classes and their spatial characteristics. Experimental results show that MMO-IG effectively generates high-quality RS images and improves detection performance.

However, MMO-IG has some limitations. When handling rare instances, the generated images may fail to accurately represent them. Function fitting sometimes oversimplifies instance attributes, and the method’s performance can degrade with large numbers of targets due to increased complexity. Additionally, while MMO-IG performs well on remote sensing images, it may struggle with natural scenes that involve more complex backgrounds, lighting variations, and occlusions. Finally, the computational requirements of the image generation process limit its application in resource-constrained environments.
We aim to address these challenges in future work by improving the framework’s adaptability, reducing its computational demands, and enhancing its handling of rare instances and large target counts.
\bibliographystyle{IEEEtran}
\bibliography{egbib}

\begin{IEEEbiography}[{\includegraphics[width=1in,height=1.25in,clip,keepaspectratio]{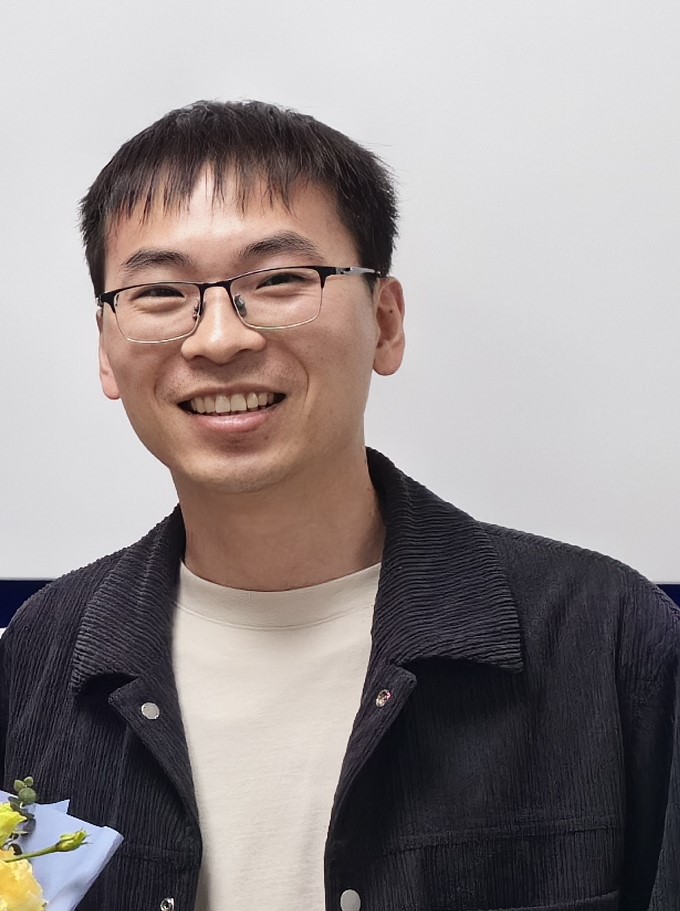}}]{Chuang Yang} received the B.E. degree in automation and the M.E. degree in control engineering from Civil Aviation University of China, Tianjin, China, in 2017 and 2020 respectively. He is currently working toward the Ph.D. degree in the School of Computer Science and School of Artificial Intelligence, OPtics and ElectroNics (iOPEN), Northwestern Polytechnical University, Xi'an, China. His research interests include computer vision, embodied AI, and intelligent transportation.
\end{IEEEbiography}

\begin{IEEEbiography}[{\includegraphics[width=1in,height=1.25in,clip,keepaspectratio]{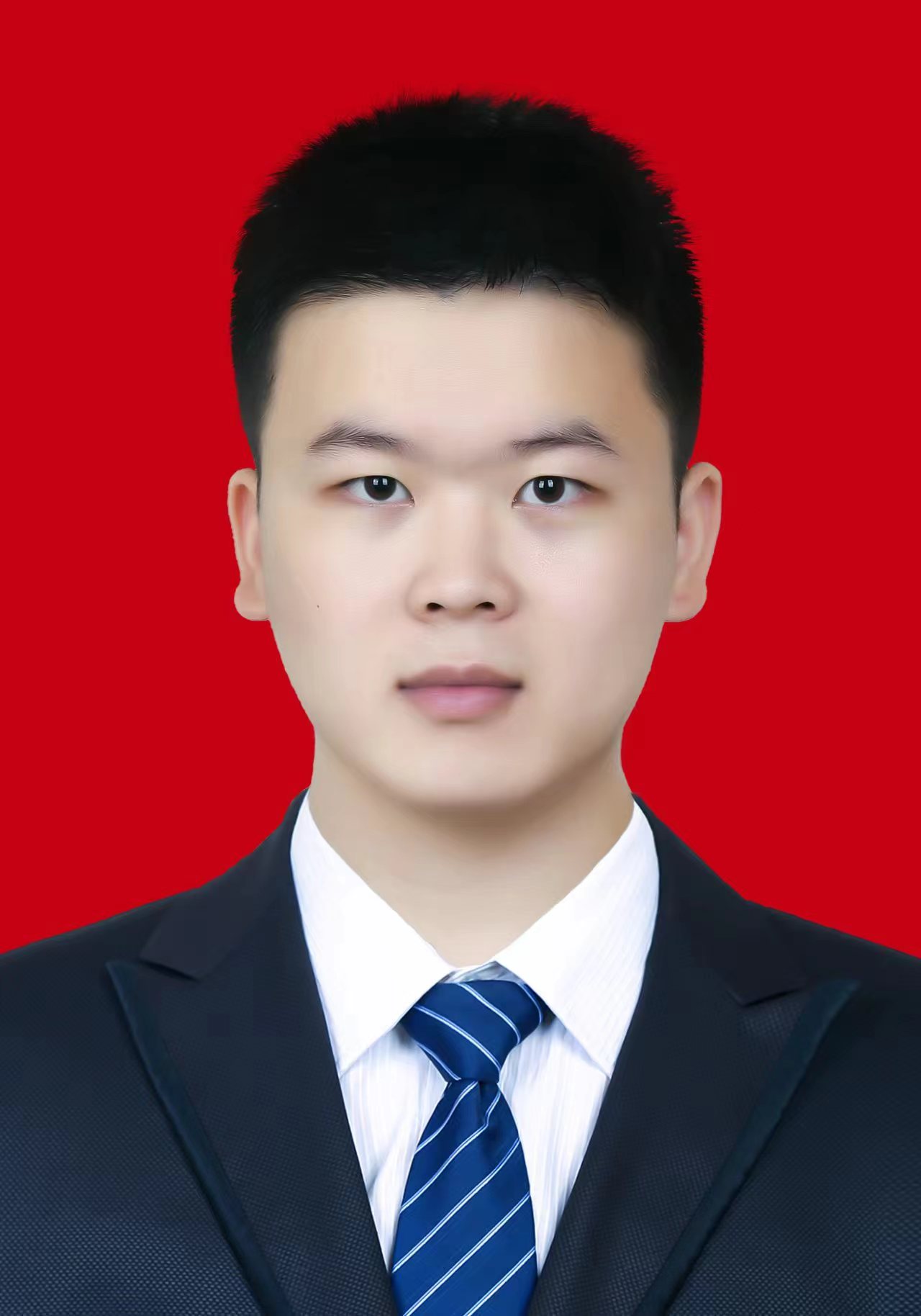}}]{Bingxuan Zhao} received the B.Sc. degree in Information and Computing Science from Northwestern Polytechnical University, Xi'an, China, in 2024. He is currently working toward the Ph.D. degree in the School of Computer Science and School of Artificial Intelligence, OPtics and ElectroNics (iOPEN), Northwestern Polytechnical University, Xi'an, China. His research interests include computer vision and machine learning.
\end{IEEEbiography}

\begin{IEEEbiography}[{\includegraphics[width=1in,height=1.25in,clip,keepaspectratio]{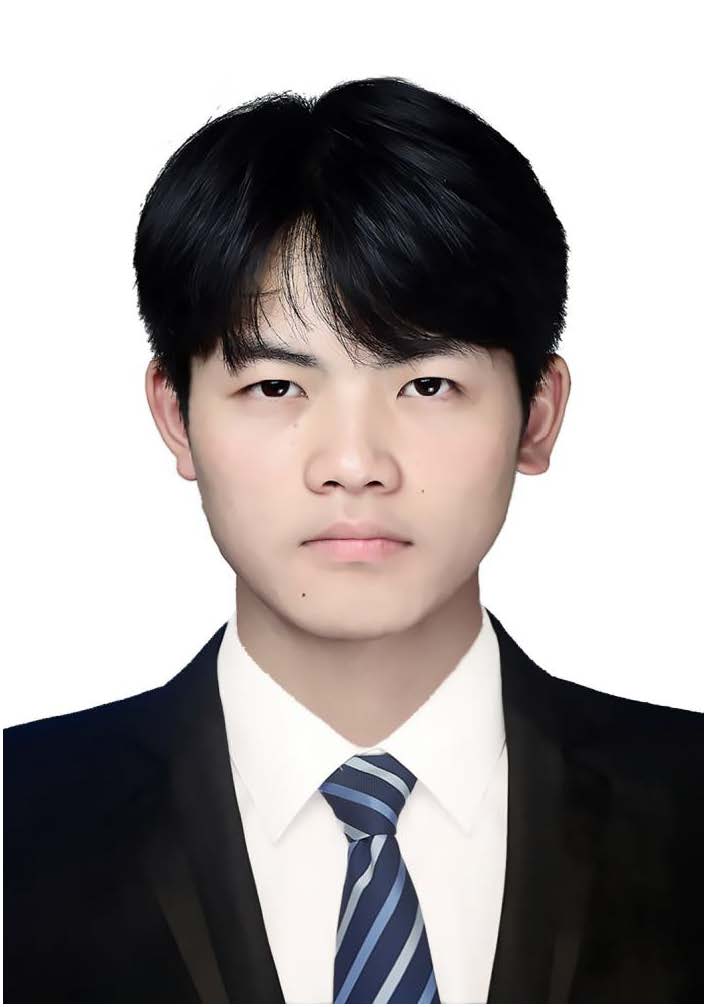}}]{Qing Zhou} is currently pursuing the Ph.D degree in computer science and technology with the school of Artificial Intelligence, Optics and Electronics (iOPEN). His research interests include computer vision and pattern recognition.
\end{IEEEbiography}

\begin{IEEEbiography}[{\includegraphics[width=1in,height=1.25in,clip,keepaspectratio]{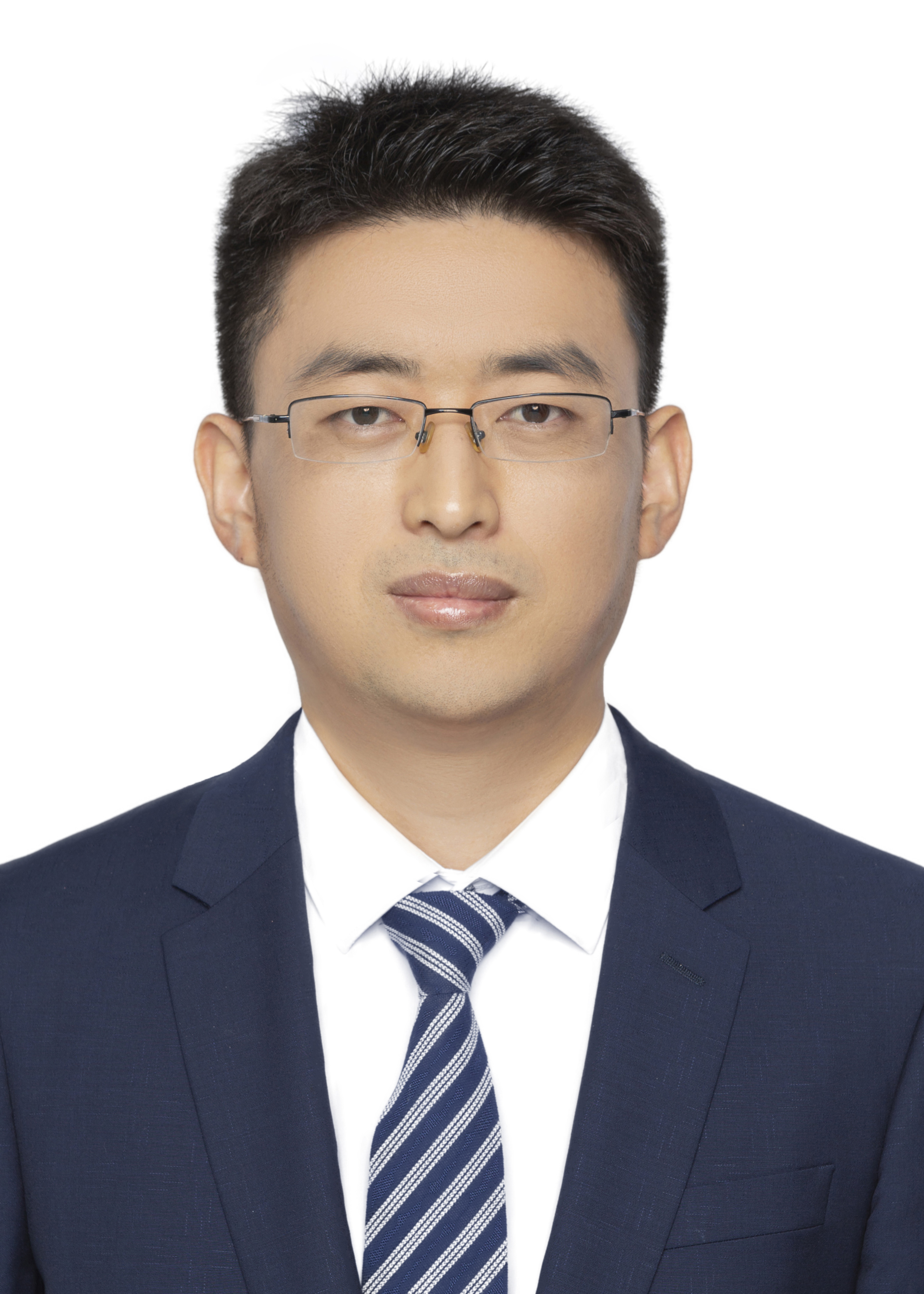}}]{Qi Wang} (M'15-SM'15) received the B.E. degree in automation and the Ph.D. degree in pattern recognition and intelligent systems from the University of Science and Technology of China, Hefei, China, in 2005 and 2010, respectively. He is currently a Professor with the School of Artificial Intelligence, OPtics and ElectroNics (iOPEN), Northwestern Polytechnical University, Xi'an, China. His research interests include computer vision, pattern recognition and remote sensing.For more information, visit the link (http://crabwq.github.io/).
\end{IEEEbiography}

\end{document}